\documentclass{article}
\usepackage[final]{neurips_2023}

\usepackage[utf8]{inputenc} 
\usepackage[T1]{fontenc}    
\usepackage{url}            
\usepackage{booktabs}       
\usepackage{amsfonts}       
\usepackage{nicefrac}       
\usepackage{microtype}      
\usepackage[table,svgnames]{xcolor}
\usepackage{printlen}
\usepackage{wrapfig}

\usepackage{enumitem}
\setlist[itemize]{leftmargin=*}

\usepackage{algorithm, algorithmic}

\definecolor{detailcolor}{RGB}{120, 120, 120}
\newcommand{\detail}[1]{{\color{detailcolor} #1}}

\definecolor{linkcolor}{RGB}{74, 102, 146}
\usepackage[
colorlinks=true,allcolors=linkcolor,pageanchor=true,
plainpages=false,pdfpagelabels,bookmarks,bookmarksnumbered,
backref=page,
]{hyperref}

\renewcommand*{\backref}[1]{}
\renewcommand*{\backrefalt}[4]{\ifcase #1 Not cited.%
  \or Cited on page~#2.%
  \else Cited on pages #2.%
  \fi%
}

\usepackage{graphicx}
\usepackage{tikz}
\usetikzlibrary{positioning}

\usepackage{caption}
\usepackage{subcaption}
\usepackage{xspace}

\usepackage{amsmath,amsfonts,bm,mathtools,amssymb}

\def\1{\bm{1}}

\def\gD{{\mathcal{D}}}

\def\gL{{\mathcal{L}}}

\DeclareMathOperator{\E}{\mathbb{E}}

\newcommand{\R}{\mathbb{R}}

\DeclareMathOperator*{\argmax}{arg\,max}
\DeclareMathOperator*{\argmin}{arg\,min}

\DeclareMathOperator*{\subjectto}{subject\;to}
\DeclareMathOperator*{\st}{s.t.}

\DeclarePairedDelimiterX{\infdivx}[2]{(}{)}{%
  #1\;\delimsize|\delimsize|\;#2%
}

\newcommand{\defeq}{\vcentcolon=}

\newcommand{\taskloss}{\ensuremath{\gL_{\text{task}}}}
\newcommand{\predloss}{\ensuremath{\gL_{\text{pred}}}}

\usepackage[nameinlink]{cleveref}
\Crefname{algorithm}{Alg.}{Algs.}

\newtheorem{theorem}{Theorem}

\newcommand{\eg}{e.g.\xspace}
\newcommand{\ie}{i.e.\xspace}
\newcommand{\cf}{\emph{cf.}\xspace}

\definecolor{lightpurple}{RGB}{168, 141, 201}

\newcommand{\cellhi}{\cellcolor{RoyalBlue!15}}
\newcommand{\cblock}[3]{
  \hspace{-1.5mm}
  \begin{tikzpicture}[node/.style={square, minimum size=10mm, thick, line width=0pt}]
    \node[fill={rgb,255:red,#1;green,#2;blue,#3}] () [] {};
  \end{tikzpicture}%
}

\title{\Large TaskMet: Task-Driven Metric Learning for Model Learning}

\author{Dishank Bansal\thanks{Work done as part of the Meta AI residency program.} \quad Ricky T.~Q.~Chen\quad
  Mustafa Mukadam\quad Brandon Amos \\
  Meta}

\begin{document}

\maketitle

\begin{abstract}
Deep learning models are often deployed in downstream tasks
that the training procedure may not be aware of.
For example, models solely trained to achieve accurate predictions
may struggle to perform well on downstream tasks because seemingly
small prediction errors may incur drastic task errors.
The standard end-to-end learning approach is to make the task
loss differentiable or to introduce a differentiable surrogate
that the model can be trained on.
In these settings, the task loss needs to be carefully balanced
with the prediction loss because they may have conflicting objectives.
We propose take the task loss signal one level deeper than
the parameters of the model and use it to learn the parameters
of the loss function the model is trained on, which can be
done by learning a metric in the prediction space.
This approach does not alter the optimal prediction model
itself, but rather changes the model learning to emphasize
the information important for the downstream task.
This enables us to achieve the best of both worlds:
a prediction model trained in the original prediction space while
also being valuable for the desired downstream task.
We validate our approach through experiments
conducted in two main settings: 1) decision-focused model learning
scenarios involving portfolio optimization and budget allocation, and
2) reinforcement learning in noisy environments with distracting states.
The source code to reproduce our experiments is available
\href{https://github.com/facebookresearch/taskmet}{here}.
\end{abstract}

\vspace*{2mm}
\section{Introduction}
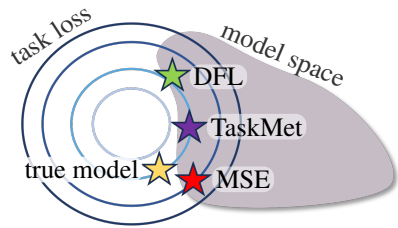
\begin{wrapfigure}{R}{2in}
    \vspace*{-7mm}
    \begin{tikzpicture}[every node/.style={anchor=south west}, inner sep=.5mm]
    \node at (0,0) {\includegraphics[width=2in]{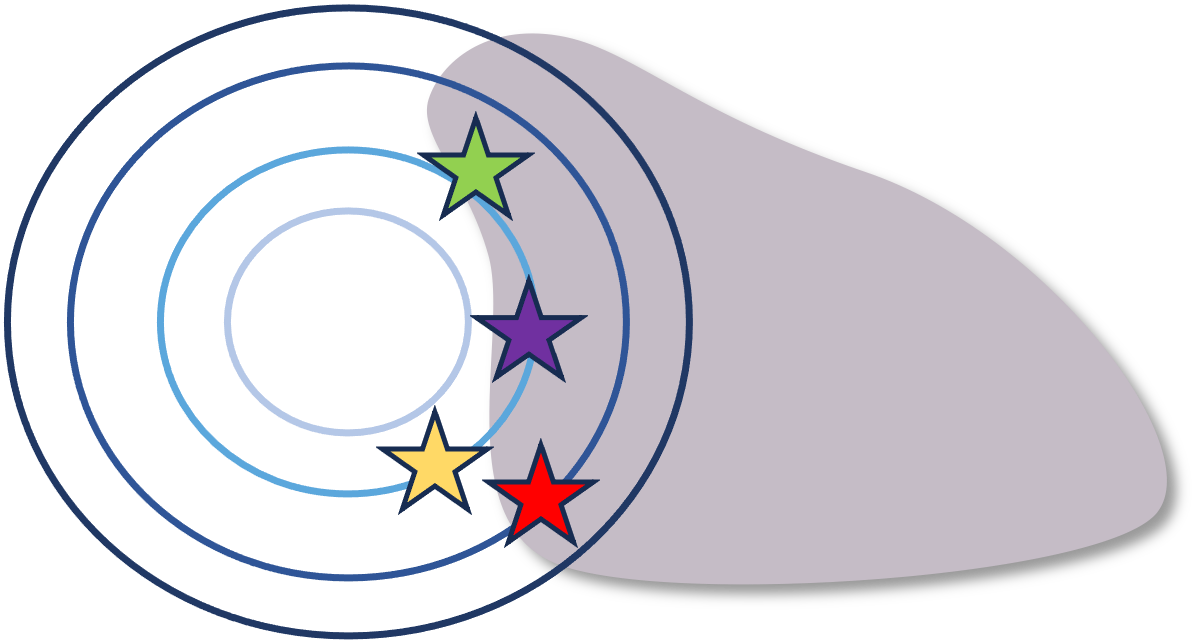}};
    \node[fill=white, fill opacity=.8, text opacity=1, rounded corners] at (0.06cm,.63cm) {true model};
    \node[fill=white, fill opacity=.5, text opacity=1, rounded corners] at (2.6cm,.5cm) {MSE};
    \node[fill=white, fill opacity=.5, text opacity=1, rounded corners] at (2.53cm,1.2cm) {TaskMet};
    \node[fill=white, fill opacity=.5, text opacity=1, rounded corners] at (2.3cm,1.87cm) {DFL};
    \node[text=black!80, rotate=-23] at (2.6cm,2.48cm) {model space};
    \node[text=black!80, rounded corners,rotate=33] at (0cm,2.1cm) {task loss};
    \end{tikzpicture}
    \vspace*{-5.5mm}
    \captionsetup{font={footnotesize}}
    \caption{\footnotesize The \emph{MSE} results in a model close to the true model
      in the prediction space, but may give poor task performance.
      \emph{Decision-focused learning} (DFL) methods optimize the task loss, but may
      deviate from the prediction space.
      \emph{TaskMet} optimizes the task loss while retaining the prediction task.
    }
    \vspace*{-5mm}
    \label{fig:models-comparison}
\end{wrapfigure}
Machine learning models for prediction are typically trained
to maximize the likelihood on a training dataset.
While the models are capable of universally approximating the
underlying data generating process to predict the output,
they are prone to approximation errors due to limited
training data and model capacity.
These errors lead to suboptimal performance in downstream
tasks where the models are used.
Furthermore, even though a model may appear to have reasonable
predictive performance on the metric and training data it
was trained on, such as the mean squared error,
employing the model for a downstream task may require the model to focus on different parts of the data
that were not emphasized in the training for predictive performance.
Overcoming the discrepancy between the model's prediction task
and performance on a downstream task is the focus of our paper.

Examples of settings where the model's prediction
loss $\predloss$ is mis-matched from the downstream task $\taskloss$ include
the following, which \cref{tab:tasks} also summarizes:
\begin{enumerate}
\item the \emph{portfolio optimization} setting from \citet{wilder2019melding},
which predicts the expected returns from stocks for a
financial portfolio. Here, the $\predloss$ is the MSE
and $\taskloss$ is from the regret of
running a portfolio optimization problem on the output;
\item the \emph{allocation} setting from
\citet{wilder2019melding},
which predicts the value of items that are being
allocated, \eg click-through-rates for recommender systems.
Here, $\predloss$ is the MSE
and $\taskloss$ measures the result of allocating the highest-value items.
\item the \emph{model-based reinforcement learning} setting of
learning the system dynamics from
\citet{nikishin2022control}.
Here, $\predloss$ is the MSE of dynamics model and the
$\taskloss$ measures how well the agent performs
for downstream value predictions.
\end{enumerate}

\begin{table}[t]
  \centering
  \caption{Settings we focus on where there is a
    discrepancy between the prediction task of a model and the downstream
    task where the model is deployed, \ie, $\predloss\neq\taskloss$.
  }
  \hspace*{-7mm}
  \resizebox{1.1\linewidth}{!}{
  \begin{tabular}{rcccc} \toprule
    & \footnotesize\detail{($x$)} & \footnotesize\detail{($y$)} &
        \footnotesize\detail{($\predloss$)} & \footnotesize\detail{($\taskloss$)} \\
    Setting & Features & Predictions & Prediction Loss & Task Loss \\ \midrule
    Portfolio optimization & Stock information & Expected return of a stock & MSE & Portfolio's performance \\
    Budget allocation & Item information & Value of item & MSE & Allocation's performance \\
    Model-based RL & Current state and action & Next state & MSE & Value estimation given the model \\
    \bottomrule
  \end{tabular}}
  \label{tab:tasks}
\end{table}

Motivated by examples such as in \cref{tab:tasks},
the research topics of
\emph{end-to-end task-based model learning} \citep{bengio1997using,donti2017task},
\emph{decision-focused learning} \citep{wilder2019melding}, and
\emph{Smart ``Predict, then Optimize''} \citep{elmachtoub2022smart}
study how to use information from the downstream task to improve the model's
performance on that particular task.
Task-based learning has applications in
financial price predictions \citep{bengio1997using,elmachtoub2022smart},
inventory stock, demand, and price forecasting \citep{donti2017task,elmachtoub2022smart,el2019generalization,mandi2020smart,liu2023active},
dynamics modeling for model-based reinforcement learning
\citep{farahmand2017value,amos2018differentiable,farahmand2018iterative,bhardwaj2020differentiable,voelcker2022value,nikishin2022control},
renewable nowcasting \citep{vohra2023end},
vehicular routing \citep{shi2023decision},
restless multi-armed bandits for maternal and child care \citep{wang2022decision},
medical resource allocation \citep{chung2022decision}, and
budget allocation, matching, and recommendation problems
\citep{kang2019few,wilder2019melding,shah2022decisionfocused}.

\textbf{Limitations of task-based learning.}
Task-based model learning comes with the goal of being able to
discover task-relevant features and data-samples on its own without
the need of explicit inductive biases.
The current trend for end-to-end model learning uses task loss along
with the prediction loss to train the prediction models.
Though easy to use, these methods may be limited by
1) the prediction overfitting to the particular task, rendering it unable to generalize;
2) the need to tuning the weight combining the task and prediction losses as in \cref{eq:dfl}.

\textbf{Our contributions.}
We propose one way of overcoming these limitations:
use the task-based learning signal not to directly optimize the weights
of the model, but to \emph{shape} a prediction
loss that is constructed in a way so that the model will always stay in
the original prediction space.
We do this in \cref{sec:learning}
via metric learning in the prediction space
and use the task signal to learn a parameterized Mahalanobis loss.
This enables more interpretable learning of the model using the metric
compared to learning with a combination of task loss and prediction loss.
The learned metric can uncover underlying properties of the task
that are useful for training the model, \eg as in
\cref{fig:learned-metric-distractors,fig:learned-metrics-quadratic}.
\Cref{sec:exps} shows the empirical success of metric learning on
decision focused model learning and model-based reinforcement learning.
\Cref{fig:models-comparison} illustrates the differences to prior methods.

\section{Background and related work}
\label{sec:prelim}
\textbf{Task-based model learning}.
We will mostly focus on solving regression problems where the
dataset $\gD \defeq \{(x_i, y_i)\}_{i=1}^N$
consists of $N$ input-output pairs, which we will assume to
be in Euclidean space.
The model makes a prediction $\hat y \defeq f_\theta(x)$
and is parameterized by $\theta$.
The model has an associated prediction loss, $\predloss$,
and is used in conjunction with some downstream task that provides
a task loss, $\taskloss$, which characterizes how well the
model performs on the task.
The most relevant related work to ours includes the approaches of
\citet{bengio1997using,donti2017task,farahmand2017value,kang2019few,wilder2019melding,nikishin2022control,shah2022decisionfocused,voelcker2022value,nikishin2022control,anonymous2023predictthenoptimize,shah2023leaving}, which learn the optimal prediction model
parameter $\theta$ to minimize the task loss $\taskloss$:
\begin{equation}
\label{eq:dfl}
  \theta^\star \defeq \argmin_\theta \taskloss(\theta) +
     \alpha \predloss(\theta),
\end{equation}
where $\alpha$ is a regularization parameter to weigh the prediction
loss which is MSE error (\cref{eq:mse}) in general.
Alternatives to \cref{eq:dfl} include
1) \emph{Smart, ``Predict, then Optimize''} (SPO) methods
\citep{elmachtoub2022smart,el2019generalization,mandi2020smart,liu2023active},
which consider surrogates for when the derivative is undefined
or uninformative, or
2) changing the prediction space from the original domain
into a latent domain with task information, \eg
task-specific latent dynamics for RL
\citep{hafner2019learning,hafner2019dream,hansen2022temporal}.
Extensions such as
\citet{gupta2023data,zharmagambetov2023landscape,ferber2023surco}
learn surrogates to overcome computationally expensive losses in \cref{eq:dfl}.
\citet{sadana2023survey} provide a further survey of this research area.

Separate from above line of work, the computer vision and NLP communities
have also considered task-based losses for models:
\citep{pinto2023tuning} tune vision models with task rewards,
\eg for detection, segmentation, colorization, and captioning;
\citet{wu2021fixes} consider representation learning for multiple tasks,
\citet{fernando2021dynamically,phan2020resolving} consider weighted loss
for class imbalance problems in classification, object detection.

Works such as \citet{farahmand2017value, voelcker2022value} use task loss in a different way compared to the above methods. They use task loss as a weighting term in the MSE loss itself. So the models are trained to focus more on samples with higher task loss. In their work, the task is the estimation of the value function in model-based RL. This can be seen as the instantiation of our work where the task loss is directly used as a metric instead of learning a metric. 

Other related work on metric learning such as
\citet{hastie1995discriminant,yang2006distance,weinberger2007metric,kulis2013metric,hauberg2012geometric,kaya2019deep}
often learns a non-Euclidean metric or distance
that captures the geometry of the data and then
solves a prediction task such as regression, clustering,
or classification in that geometry.
Other methods such as
\citet{voelcker2022value}
can handcraft metrics based on domain knowledge. 
In contrast to these, in the task-based model learning,
we propose that the downstream task (instead of the data alone)
gives the relevant metric for the prediction,
and that it is possible to use end-to-end learning
as in \cref{eq:method_overall} to obtain the task-based metric.

\textbf{How our contribution fits in.}
The mentioned methods mainly deal with using task-based losses to condition the model learning either by differentiation through task loss to update the model or using it directly as weighing for MSE prediction loss. Whereas our work focuses on using task loss to \emph{learn} a parameterized prediction loss which is then used to train the model. The task loss is not \emph{directly} used for model training.


\section{Task-driven metric learning for model learning}
\label{sec:learning}

We first present why it's useful to see the prediction space as a
non-Euclidean metric space with an unknown metric, then show
how task-based learning methods can be used to learn that metric.

\subsection{Metrics in the prediction space --- Mahalanobis losses}
When defining a loss on the model, we are forced to make a choice
about the geometry to quantify how good a prediction is.
This geometric information is often implicitly set in standard
learning settings and there are often no other reasonable choices without more information.
For example, a supervised model $f_\theta$ parameterized by $\theta$
is often trained with the mean squared error (MSE)
\begin{equation}
\theta^\star_\text{MSE} \defeq \argmin_\theta \E_{(x,y) \sim \gD}\left[(f_\theta(x)-y)^2\right].
\label{eq:mse}
\end{equation}
The MSE makes the assumption that the geometry of the prediction
space is Euclidean.
While it is a natural choice, it may not be ideal when
the model needs to focus on important parts of the data
that are under-emphasized under the Euclidean metric.
This could come up by needing to emphasize some samples over
others, or some dimensions of the prediction space over others.

While there are many possible geometries and metric spaces that could
be defined over prediction spaces, they are difficult to specify
without more information.
We focus on the metric space defined by the Mahalanobis norm
$\|z\|_M\defeq \left(z^\top M z\right)^{1/2}$, where
$M$ is a positive semi-definite matrix.
The Mahalanobis norm results in the prediction loss
\begin{equation}
\label{eq:implicit_theta}
\gL_{\text{pred}}(\theta, \phi) \defeq
\E_{(x,y) \sim \gD}\left[\|f_\theta(x)-y\|_{\Lambda_\phi(x)}^2\right],
\end{equation}
where $\Lambda_\phi$ is a metric parameterized by
$\phi$ and this is also conditional on the feature $x$ so it can
learn the importance of the regression space from each
part of the feature space.

Many methods can be seen as hand-crafted ways of setting a Mahalanobis metric,
including:
1) normalizing the input data by making the metric appropriately scale
the dimensions of the prediction,
2) re-weighting the samples as in \citet{donti2017task,lambert2020objective}
by making the metric scale each sample based on some importance factor,
or
3) using other performance measures, such as the value gradient in \citet{voelcker2022value}.

\begin{figure}[t]
  \centering
  \begin{tikzpicture}
    \node[align=left,anchor=north west] (losses) at (0,0) {\includegraphics{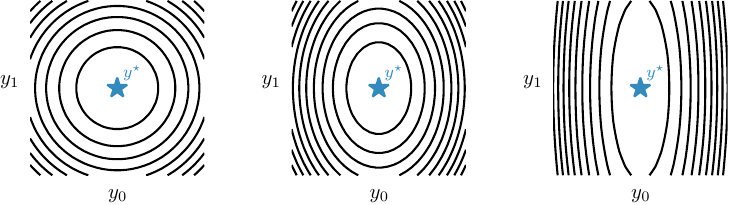}};
    \node at (.cm,.3cm) {$\Lambda_\phi(x)$};
    \node at (2.1cm,.5cm) {
      \begin{minipage}{2cm}
        $$\begin{bmatrix}
            1 & 0 \\
            0 & 1
          \end{bmatrix}$$
      \end{minipage}};
    \node at (6.55cm,.5cm) {
      \begin{minipage}{2cm}
        $$\begin{bmatrix}
            2 & 0 \\
            0 & 1
          \end{bmatrix}$$
      \end{minipage}};
    \node at (10.95cm,.5cm) {
      \begin{minipage}{2cm}
        $$\begin{bmatrix}
            10 & 0 \\
            0 & 1
          \end{bmatrix}$$
      \end{minipage}};
  \end{tikzpicture}
  \caption{Examples of the Mahalanobis loss from
    \cref{eq:implicit_theta} in a 2-dimensional prediction task.
    The model's loss is zero only when $\hat y=y^\star$.
    Here, the metric $\Lambda_\phi(x)$ increases the weighting
    on the $y_0$ component of the loss and thus emphasizes
    the predictions along this dimension.
  }
  \label{fig:mahalanobis-loss-ex}
\end{figure}
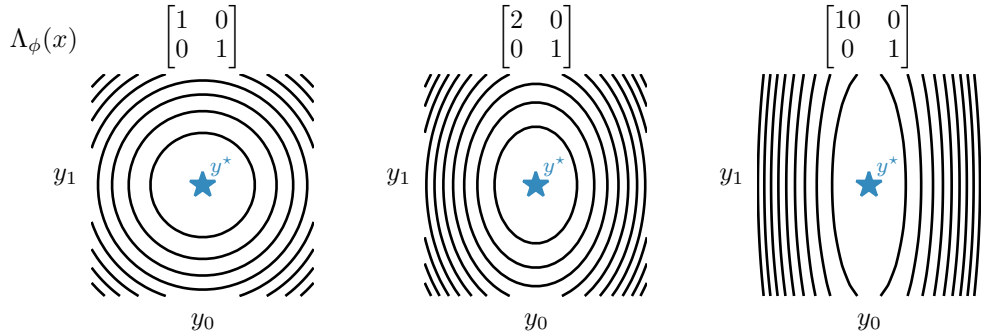

More generally beyond these, the Mahalanobis metrics help emphasize the:
\begin{enumerate}
\item \emph{relative importance of dimensions}.
  the metric allows for down- or up-weighting different dimensions of
  the prediction space by changing the diagonal entries of the metric.
  \Cref{fig:mahalanobis-loss-ex} illustrates this.
\item \emph{correlations in the prediction space}.
  the quadratic nature of the loss with the metric allows the model to
  be aware of correlations between dimensions in the prediction space.
\item \emph{relative importance of samples}.
  heteroscedastic metrics $\Lambda(x)$ enable different samples to be
weighted differently for the final expected cost over the dataset.
\end{enumerate}

Without more information, parameterizing and specifying the best metric
for learning the model is challenging as it involves the subproblem
of understanding the relative importance between predictions.
We suggest that when it is available, the downstream task information
characterizing the overall model's performance can be used to learn
a metric in the prediction space.
Hence, learning model parameters with a metricized loss can be seen as
conditioning the learning problem. The ability to learn the metric
end-to-end enables the task to condition the learning of the model in
any or all of the three ways described above. This approach offers an
interpretable method for the task to guide the model learning, in
contrast to relying solely on task gradients for learning model
parameters, which may or may not align effectively with the given
prediction task.

\subsection{End-to-end metric learning for model learning}
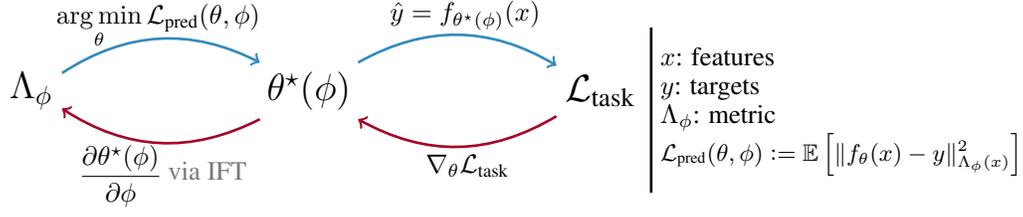
\begin{figure}[t]
  \centering
  \definecolor{bmh_blue}{HTML}{348ABD}
  \definecolor{bmh_red}{HTML}{A60628}
  \hspace*{-19mm}
  \begin{tikzpicture}
    \node (metric) at (0,0) {\Large $\Lambda_\phi$};
    \node[right=2.6cm of metric] (params) {\Large $\theta^\star(\phi)$};
    \node[right=2.6cm of params] (taskloss) {\Large $\taskloss$};
    \node at (1.69cm,1.05cm) {
      \begin{minipage}{3cm}
      $$\argmin_\theta \predloss(\theta, \phi)$$
      \end{minipage}};
    \node at (5.77cm,1.0cm) {$\hat y=f_{\theta^\star(\phi)}(x)$};
    \node at (1.75cm,-1.1cm) {
      \begin{minipage}{1.1cm}
      $$\frac{\partial \theta^\star(\phi)}{\partial \phi}$$
      \vspace{-1mm}
    \end{minipage}
    \detail{via IFT}
  };
    \node at (5.8cm,-1.cm) {$\nabla_{\theta}\taskloss$};

    \draw[->,draw=bmh_blue, line width=1] (metric) edge[bend right=-30] node {} (params);
    \draw[->,draw=bmh_blue, line width=1]  (params) edge[bend right=-30] node {} (taskloss);
    \draw[->,draw=bmh_red, line width=1] (taskloss) edge[bend right=-30] node {} (params);
    \draw[->,draw=bmh_red, line width=1] (params) edge[bend right=-30] node {} (metric);
  \end{tikzpicture} \hspace{-2mm}
  \begin{tikzpicture}
    \node[align=left,anchor=south west] (taskloss) at (0, 0) {
      $x$: features \\
      $y$: targets \\
      $\Lambda_\phi$: metric \\
      {\footnotesize $\predloss(\theta, \phi)\defeq \E\left[\|f_\theta(x)-y\|_{\Lambda_\phi(x)}^2\right]$}
    };
    \draw[-, line width=1pt, draw=black] (0,-.1) edge node {} (0,2.1);
    \node at (0,-0.5cm) {}; 
  \end{tikzpicture}\hspace*{-20mm}
  \caption{TaskMet learns a metric for predictions with the gradient
    from a downstream task loss.
  }
  \label{fig:method}
\end{figure}

Our key idea is to learn a metric
in the form of \cref{eq:implicit_theta}
end-to-end with a given task, which is then used to train the prediction model.
\Cref{fig:method,alg:taskmet} summarize this approach.
The learning problem of the metric and model parameters are
formulated as the bilevel optimization problem
\begin{align}
\label{eq:method_overall}
\phi^\star \defeq \argmin_\phi &\;\; \taskloss(\theta^\star(\phi)), \\
\label{eq:model-implicit}
\subjectto &\;\; \theta^\star(\phi) = \argmin_\theta \predloss(\theta, \phi)
\end{align}
where $\phi$ and $\theta$ are \detail{(respectively)} the metric
and model parameters,
$\predloss$ is the metricized prediction loss (\cref{eq:implicit_theta}) to train
the prediction model, and
$\taskloss$ is the task loss defined by the
task at hand (which could be another optimization problem,
\eg \cref{eq:lodl-task}, or another learning task,
\eg \cref{eq:omd-task}.

\textbf{Gradient-based learning.}
We learn the optimal metric $\Lambda_{\phi^\star}$ with the
gradient of the task loss, \ie $\nabla_\phi \taskloss(\theta^\star(\phi))$.
Using the chain rule and assuming we have the optimal $\theta^\star(\phi)$
for some metric parameterization $\phi$, this derivative is
\begin{equation}
\label{eq:dLtask_dphi}
    \nabla_\phi\taskloss(\theta^\star(\phi)) = \nabla_\theta\taskloss(\theta)\big|_{\theta=\theta^\star(\phi)} \cdot \frac{\partial \theta^\star(\phi)}{\partial\phi}
\end{equation}



To calculate the term $\nabla_\phi\taskloss(\theta^\star(\phi))$, we
need to compute two gradient terms:
$\nabla_\theta\taskloss(\theta)\big|_{\theta=\theta^\star(\phi)}$ and
$\partial \theta^\star(\phi)/\partial\phi$. The former can be
estimated in standard way since $\taskloss(\theta)$ is an explicit
function of $\theta$. However, the latter cannot be directly
calculated because $\theta^\star$ is a function of optimization
problem which is multiple iterations of gradient descent, as shown in
\cref{eq:model-implicit}. Backpropping through multiple iterations of
gradient descent can be computationally expensive, so
we use the implicit function theorem
(\cref{sec:ift}) on the first-order
optimality condition of \cref{eq:model-implicit}, \ie
$\frac{\partial\predloss(\theta, \phi)}{\partial \theta} = 0$.
Combining these,
$\nabla_\phi\taskloss(\theta^\star(\phi))$ can be computed with
\begin{equation}
  \label{eq:grad-phi}
  \nabla_\phi\taskloss(\theta^\star(\phi)) =
  \nabla_\theta\taskloss(\theta)  \cdot
  \underbrace{-\left(\frac{\partial^2 \predloss(\theta, \phi)}{\partial \theta^2}\right)^{-1}
  \frac{\partial^2 \predloss(\theta, \phi)}{\partial\theta \partial\phi}\Bigg|_{\theta=\theta^\star(\phi)}}_{\partial \theta^\star / \partial\phi}
  \hspace*{-5mm}
\end{equation}

The implicit derivatives in \cref{eq:grad-phi} may be challenging to compute
or store in memory because the Hessian term
${\partial^2 \predloss(\theta, \phi)}/{\partial \theta^2}$
is the Hessian of the prediction loss with respect to the model's
parameters.
Approaches such as \citet{lorraine2020optimizing} are able to
scale related implicit differentiation problems to models with
millions of hyper-parameters.
The main insight is that the Hessian does not need to be
explicitly formed or inverted and the entire implicit derivative
term needed for backpropagation can be obtained with
an implicit solver.
We follow \citet{blondel2022efficient} and compute the
implicit derivative by using conjugate gradient on the
normal equations.

\begin{algorithm}[t]
  \caption{TaskMet: Task-Driven Metric Learning for Model Learning}
  \begin{algorithmic}
    \STATE \textbf{Models:} predictor $f_\theta$ and metric $\Lambda_\phi$
    with initial parameterizations $\theta$ and $\phi$
    \WHILE{unconverged}
    \STATE \emph{// approximate $\theta^\star(\phi)$ given the current metric $\Lambda_\phi$}
    \FOR{$i$ in $1\dots K$}
    \STATE $\theta \leftarrow \text{update}(\theta, \nabla_\theta \predloss(\theta, \phi))$ \emph{//
    fit the predictor $f_\theta$ to the current metric loss} {\footnotesize(\cref{eq:implicit_theta})}
    \ENDFOR
    \STATE $\phi \leftarrow \text{update}(\phi, \nabla_\phi \taskloss)$ \emph{// update the metric $\Lambda_\phi$ with the task loss} {\footnotesize(\cref{eq:dLtask_dphi})}
    \ENDWHILE
    \STATE \textbf{return} optimal predictor $f_\theta$ and metric $\Lambda_\phi$
    solving the bi-level problem in \cref{eq:method_overall}
  \end{algorithmic}
  \label{alg:taskmet}
\end{algorithm}

\section{Experiments}
\label{sec:exps}

We evaluate our method in two distinct settings: 1) when the
downstream task involves an optimization problem parameterized by the
prediction model output, and 2) when the downstream task is another
learning task. For the first setting, we establish our baselines by
replicating experiments from previous works such as
\citet{shah2022decisionfocused} and \citet{wilder2019melding}. These
baselines encompass tasks like portfolio optimization and budget
allocation. In the second setting, we focus on model-based
reinforcement learning. Specifically, we concentrate on learning a
dynamics model (prediction model) and aim to optimize the Q-value
network using the learned dynamics model for the Cartpole task
\citep{nikishin2022control}.
\Cref{sec:impl-details} provides further experimental details
and hyper-parameter information.

\subsection{Metric parameterization}
We parameterize the metric using a neural network with parameters
$\phi$, denoted as $\Lambda_\phi \defeq L_\phi^{\top}L_\phi$, where
$L_\phi$ is an $n \times n$ matrix, where $n$ is the dimension of the
prediction space. This particular factorization constraint ensures
that the matrix is positive semi-definite, which is crucial for
it to be considered a valid metric. The neural network
parameters are initialized to make $\Lambda_\phi$ closer to the
identity matrix $\mathbb{I}$, representing the Euclidean metric. The
learned metric can be conditional on the input, denoted as
$\Lambda_\phi(x)$, or unconditional, represented as $\Lambda_\phi$,
depending on the problem's structure.

\subsection{Decision-Focused Learning}
\label{sec:lodl-exp}

\subsubsection{Background and experimental setup}
We use three standard resource allocation tasks
for comparing task-based learning methods
\citep{shah2022decisionfocused, wilder2019melding, donti2017task,
  futoma2020popcorn}. In this setting, resource utility prediction based on
some input features constitute a prediction model, resource allocation
constitutes the downstream task which is characterized by $\taskloss$
The prediction model's output parameterized the downstream
resource optimization. The settings are implemented
exactly as in \citet{shah2022decisionfocused} and have
task losses defined by
\begin{equation}
  \label{eq:lodl-task}
  \taskloss(\theta) \defeq \E_{(x,y) \sim \gD} [g(z^\star(\hat y), y)]
\end{equation}
where $z^\star(\hat y) \defeq \argmin_z g(z, \hat y)$ and
$g(z,y')$ is some combinatorial optimization objective over variable
$z$ parameterized by $y'$. The task loss $\taskloss$ is the expected value of
objective function with decision variable $z^\star(\hat y)$ induced by
the prediction model $ \hat y = f_\theta(x)$ under the ground truth
parameters $y$. We use corresponding
surrogate losses to replicate the $z^\star(\hat y)$ optimization
problem as in \citet{shah2022decisionfocused, wilder2019melding,
  xie2020differentiable} and differentiate
through the surrogate using
\verb!cvxpylayers! \citep{agrawal2019differentiable}.

These settings evaluate the ability of TaskMet to capture the
correlation between model predictions and differentiate between
different data-points in accordance to their importance for the
optimization problem. Hence, we consider a heteroscedastic metric,
\ie, $\Lambda_\phi(x)$.

\textbf{Baselines.} We compare with standard baseline losses for learning models:
\begin{enumerate}
\item The standard MSE loss $\theta^\star = \argmin_\theta \E_{(x,y) \sim \gD}[(f_\theta(x)-y)^2]$.
  This method doesn't use any task information.
\item DFL \citep{wilder2019melding}, which trains the prediction model 
  with a weighted combination of $\taskloss$ and $\predloss$ as in \cref{eq:dfl}.
\item LODL \citet{shah2022decisionfocused}, which learns a
  parametric loss for each point in the training data to approximate the
  $\taskloss$ around that point. That is, $LODL_{\psi_n}(\hat y_n) \approx
  \taskloss(\hat y_n)$ for all $n$.
  They create a dataset of $\{(\hat y_n, \taskloss(\hat y_n))\}$ for $\hat y_n$
  sampled around the $y_n$. After this they learn the LODL loss for each point
  as
  $\psi^\star_n = \argmin_{\psi_n} \frac{1}{K}\sum_{k=1}^K
  (LODL_{\psi_n}(y_n^k) - \taskloss(y_n^k))^2$.
  We chose the ``Quadratic'' variant of their method which is the closest to ours,
  where $LODL_{\psi_n}(\hat y) = (\hat y - y)^{\top}\psi_n(\hat y - y)$ where
  $\psi_n$ is a learned symmetric Positive semidefinite (PSD) matrix.
  LODL also uses \cref{eq:dfl} to learn the model parameters, but using
  $LODL_{\psi_n}(\hat y_n) \approx \taskloss(\hat y_n)$
\end{enumerate}

\textbf{Experimental settings.}
We use the following experimental settings from
\citep{shah2022decisionfocused}:
\begin{enumerate}
    \item \textbf{Cubic}: This setting evaluates methods under
model mismatch scenario where the model being learned suffers with
severe approximation error. In this task, it is important for methods to allocate model capacity to the points more critical for the downstream tasks. \\
    \emph{Prediction Model}: A linear prediction model $f_\theta(x) \defeq \theta x$ is learned for the problem where the ground truth data is generated by cubic function, \ie, $y_i = 10x_i^3 -6.5x_i, x_i \in U[-1,1]$. \\
    \emph{Downstream task}: Choose top $B=1$ out of $M=50$ resources $\hat{\mathbf{y}} = [\hat y_1, \dots,\hat y_M]$, $z^\star(\hat{\textbf{y}}) \defeq \argmax_i \hat{\mathbf{y}}$
  \item \textbf{Budget Allocation}:
    Choose top $B=2$ websites to advertise based on
    Click-through-rates (CTRs) predictions of $K$ users on $M$ websites. \\
    \emph{Prediction Model}:  $\hat{\mathbf{y}}_m = f_\theta(x_m)$ where $x_m$ is given features of $m^{\text{th}}$ website and $\hat{\mathbf{y}}_m = [\hat y_{m,1}, \dots, \hat y_{m,K}]$ is the predicted CTRs for $m^{\text{th}}$ website for all $K$ users. \\
    \emph{Downstream task}: Determine $B=2$ websites such that the expected number of users that click on the ad at least once is maximized, \ie, $z^\star(\hat{\mathbf{y}}_m) = \argmax_z \sum_{j=0}^K(1-\prod_{i=0}^M z_i \cdot \hat y_{ij})$ where $z_i \in \{0,1\}$.
  \item \textbf{Portfolio Optimization}:
    The task is to choose a  distribution over $M$ stocks in Markowitz portfolio optimization \citep{markowitz2000mean, michaud1989markowitz} that maximizes the expected return under the risk penalty. \\
    \emph{Prediction Model}: Given the historical data $x_m$ about a stock $m$, predict the future stock price $\hat y_m$. Combining prediction over $M$ stocks to get $\hat{\mathbf{y}} = [\hat y_1, \dots, \hat y_M]$. \\
    \emph{Downstream Task}: Given the correlation matrix $Q$ of the stocks, choose a distribution over stocks $\mathbf{z}^\star(\hat{\mathbf{y}}) = \argmax_\mathbf{z} \mathbf{z}^{\top}\hat{\mathbf{y}} - \lambda \mathbf{z}^{\top}Q\mathbf{z} \st \sum_{i=0}^M z_i \leq 1 \quad \text{and} \quad 0\leq z_i \leq 1,\forall i$
\end{enumerate}

We run our own experiments for LODL
\citep{shah2022decisionfocused} using their public code.

\begin{wrapfigure}{R}{2.3in}
  \centering
  \hspace{8mm}\colorbox{white}{\small \cblock{52}{138}{189} TaskMet \hspace{2mm} \cblock{166}{6}{40} MSE} \\
  \includegraphics[width=\linewidth]{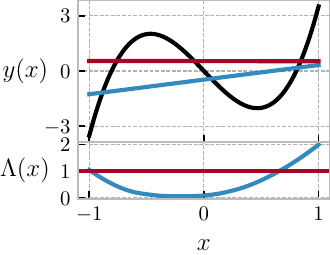} \\
  \caption{\detail{(Cubic problem)} TaskMet learns a metric that prioritizes points that are
    the most important the downstream task. The euclidean metric (MSE)
    puts equal weight on all points and leads to a bad
    model with respect to the downstream task.}
  \label{fig:learned-metrics-quadratic}
\end{wrapfigure}

\subsubsection{Experimental results}
\Cref{tab:lodl-dq} presents a summary of the
performance of different methods on all the tasks. Each problem poses
unique challenges for the methods. The \emph{cubic} setting suffers
from severe approximation errors, hence the learning method needs to
allocate limited model capacity more towards higher utility points compared to
lower utility points. The MSE method performs the worst as it lacks
task information and only care about prediction error. DFL with
$\alpha = 0$ performs better than MSE, but it can get trapped in local
optima, leading to higher variance in the problem
\citep{shah2022decisionfocused}. LODL ($\alpha=0$) performs among the
highest in this problem since it uses learned loss
for each point. TaskMet performs as good as LODL as it can capture the
relative importance of higher utility points versus lower utility
points using the learned metric, resulting in more accurate
predictions for those points (see
\cref{fig:learned-metrics-quadratic}). In \emph{budget allocation},
DFL (with $\alpha=0$) performs the best, since it is solely optimizer
over $\taskloss$, but on the other hand it has $10$ orders of larger
prediction error as shown in \cref{tab:lodl-mse} indicating that the
model is overfit to the task, LODL ($\alpha=0$) suffers
from the same problem. TaskMet has the $2^{\text{nd}}$ best Decision
Quality without overfitting on the task, \ie, low prediction error. In
\emph{Portfolio Optimization}, the decision quality correlates highly
with the model accuracy/prediction error as in this setting the
optimization problem mostly depends upon the accurate prediction of
the stocks. This is the reason that MSE, DFL ($\alpha = 10$) performs
the best, but DFL ($\alpha=0$) performs the worst, since it has solely
being trained on $\taskloss$ without any $\predloss$. As shown in
\cref{tab:lodl-dq} and \cref{tab:lodl-mse}, TaskMet is the only method
that consistently performs well considering both task loss and prediction loss, across all the problem settings, this
is due to the ability of the metric to infer problem-specific features
without manual tuning, unlike other methods.

\begin{figure}[t]
\begin{minipage}[t]{0.46\textwidth}
\begin{table}[H]
  \caption{Normalized test decision quality (DQ)
    on the decision oriented learning problems.}
  \label{tab:lodl-dq}
  \centering
  \newcommand{\entry}[2]{$#1$\detail{{\footnotesize $\pm #2$}}}
  \setlength{\tabcolsep}{4pt}
  \resizebox{\linewidth}{!}{
  \begin{tabular}{lclll}
    \toprule
    & & \multicolumn{3}{c}{Problems}                   \\
    \cmidrule(r){3-5}
    Method   & $\alpha$       &   Cubic       &   Budget &   Portfolio \\
    \midrule
    MSE &  & \entry{-0.96}{0.02}  & \entry{0.54}{0.17}       & \cellhi \entry{0.33}{0.03}   \\
    DFL & $0$   & \entry{0.61}{0.74}     & \cellhi \entry{0.91}{0.06}       & \entry{0.25}{0.02}   \\
    DFL & $10$ & \entry{0.62}{0.74} & \entry{0.81}{0.11}   & \cellhi \entry{0.34}{0.03} \\
    LODL & $0$ & \cellhi \entry{0.96}{0.005}  &  \entry{0.84}{0.105}  & \entry{0.17}{0.05}  \\
    LODL & $10$ &  \entry{-0.95}{0.005} & \entry{0.58}{0.14} & \cellhi \entry{0.30}{0.03} \\ \midrule
    TaskMet &  & \cellhi \entry{0.96}{0.005}     & \entry{0.83}{0.12}  & \cellhi \entry{0.33}{0.03}   \\
    \bottomrule
  \end{tabular}}
\detail{\footnotesize{$0$=random model $1$=oracle model}}
\end{table}
\end{minipage}
\hfill
\begin{minipage}[t]{0.50\textwidth}
\begin{table}[H]
  \caption{Test prediction errors (MSE) on the decision oriented learning problems.}
  \label{tab:lodl-mse}
  \centering
  \newcommand{\entry}[2]{$#1$\detail{{\footnotesize $\pm #2$}}}
  \setlength{\tabcolsep}{4pt}
    \resizebox{\linewidth}{!}{
  \begin{tabular}{lcccc}
    \toprule
    & & \multicolumn{3}{c}{Problems}                   \\
    \cmidrule(r){3-5}
    Method  & $\alpha$ &   Cubic  &  Budget  {\scriptsize($\times 1e^{-4}$)}   &   Portfolio  {\scriptsize($\times 1e^{-4}$)}   \\
    \midrule
    MSE  & & \entry{2.30}{0.03} & \entry{4.32}{2.35} & \entry{4.03}{0.24}\\
    DFL & 0 & \entry{2.89}{0.32}& \entry{71.7}{58.3} &	\entry{8.0e^{3}}{8e^{2}}\\
    DFL & 10 & \entry{2.41}{0.05}&\entry{8.09}{12.1}&	\entry{5.18}{0.46} \\
    LODL & 0 & \entry{2.88}{0.03} & \entry{35.9}{12.9} & \entry{55.6}{9.95} \\
    LODL & 10 & \entry{2.29}{0.19}& \entry{5.05}{1.88} & \entry{4.31}{0.31}\\ \midrule
    TaskMet  & & \entry{2.89}{0.03} & \entry{9.74}{13.79} & \entry{4.69}{0.56}\\
    \bottomrule
  \end{tabular}}
 \detail{\footnotesize{$\alpha$ is the prediction loss weight in \cref{eq:dfl}}}
\end{table}
\end{minipage}
\end{figure}

\subsection{Model Based Reinforcement Learning}
\begin{figure}[t]
  \centering
  \includegraphics[width=0.9\textwidth]{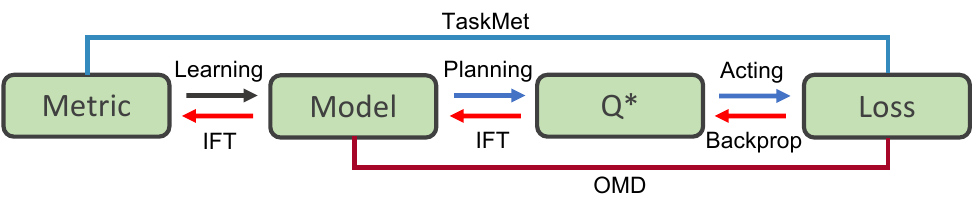}
  \caption{OMD \citep{nikishin2022control} uses the planning task loss to
    learn the model parameters using implicit gradients. TaskMet add one
    more optimization step over OMD and instead of learning the model
    parameters using task loss, we learn the metric which then is used to
    learn model parameters.}
  \label{fig:omd-vs-taskmet}
\end{figure}

\subsubsection{Background and experimental setup}
Model-based RL suffers from objective-mismatch \citep{bansal2017goal,lambert2020objective}. This is because dynamics models trained for data likelihood maximization do not translate to optimal policy. To reduce objective-mismatch, different losses \citep{farahmand2017value, voelcker2022value} have been proposed to learn the model which is better suited to learning optimal policies. TaskMet provides an alternative approach towards reducing objective-mismatch, as the prediction loss is directly learnt using task loss. 
We set up the MBRL problem as follows. Given the current state $s_t$ and control $a_t$ at a timestep $t$
of a discrete-time MDP, the \emph{dynamics model} predicts the next state transition,
\ie $\hat s_{t+1}\defeq f_\theta(s_t, a_t)$.
The prediction loss is commonly the squared error loss
$\E_{s_t, a_t, s_{t+1}}\|s_{t+1}-f_\theta(s_t, a_t)\|_2^2$,
and the downstream task is to find the optimal Q-value/policy.
\citet{nikishin2022control} introduces idea of \emph{optimal model design} (OMD)
to learn the dynamics model end-to-end with the policy objective
via implicit differentiation.
Let $Q_\omega(s,a)$ be the action-conditional value function parameterized by $\omega$.
The Q network is trained to minimize the Bellman error induced by the model $f_\theta$:
\begin{equation}
  \label{eq:q-loss}
  \gL_Q(\omega, \theta) \defeq \E_{s,a}[Q_w(s,a)-\mathrm{B}^{\theta} Q_{\bar w}(s,a)]^2,
\end{equation}
where
$\bar \omega$ is moving average of $\omega$ and $\mathrm{B}^\theta$ is the
model-induced Bellman operator
$\mathrm{B}^{\theta} Q_{\bar w}(s,a) \defeq r_\theta(s,a) +
\gamma\E_{p_\theta(s,a,s')} [\log\sum_{a'}\exp Q(s',a')]$.
Q-network optimality defines $\omega$ as an implicit function of the model
parameters $\theta$ as $\omega^\star(\theta) = \argmin_\omega
\mathcal{L}_Q(\omega, \theta) \implies \frac{\partial
\mathcal{L}_Q(\omega, \theta)}{\partial \omega} = 0$.
Now we have task loss which is optimized to find optimal Q-values:
\begin{equation}
  \label{eq:omd-task}
  \taskloss(\omega^\star(\theta)) \defeq
  \E_{s,a}[Q_{\omega^\star(\theta)}(s,a)-  \mathrm{B}Q_{\bar \omega}(s,a)]^2
\end{equation}
where the Bellman operator
induced by ground-truth trajectory and reward is
$\mathrm{B}Q_{\bar\omega}(s,a) \defeq r(s,a) + \gamma
  \E_{s,a,s'}\log \sum_{a'}\exp Q_{\bar\omega}(s',a')$.


\textbf{OMD setup.}
OMD end-to-end optimizes the model for the task loss,
\ie
$\theta^\star = \argmin_\theta \taskloss(\omega^\star(\theta))$.

\textbf{TaskMet setup.}
For metric learning, we extend OMD to learn a metric using
task gradients, to train the model parameters,
see \cref{fig:omd-vs-taskmet}.
Metric learning just adds one more level of optimization to
OMD and results in the \emph{tri-level} problem
\begin{equation}
\label{eq:taskmet-rl}
\footnotesize
\begin{aligned}
    \phi^\star = \argmin_\phi\;\; &\taskloss(\omega^\star) \\
    \subjectto\;\; & \omega^\star(\theta^\star) = \argmin_\omega \gL_Q(\omega, \theta^\star) \\
    & \theta^\star(\phi) = \argmin_\theta \predloss(\phi, \theta)
\end{aligned}
\vspace*{-3mm}
\end{equation}
where $\taskloss(\omega^\star)$ and $\gL_Q(\omega, \theta^\star)$ are
defined in \cref{eq:omd-task} and \cref{eq:q-loss}, respectively, and
$\predloss(\phi, \theta) = \E_{s_t, a_t, s_{t+1}}\|s_{t+1}-f_\theta(s_t, a_t)\|_{\Lambda_\phi(s_t)}^2$ is the metricized prediction loss in
\cref{eq:implicit_theta}.

To learn $\phi^{\star}$ using gradient descent, we estimate $\nabla_\phi \taskloss$ as
\begin{equation}
  \begin{aligned}
\nabla_\phi \taskloss =& \nabla_\omega\taskloss(\omega^\star) \cdot \frac{\partial \omega^\star}{\partial \theta^\star} \cdot \frac{\partial \theta^\star}{\partial \phi} \\
=& \nabla_\omega\taskloss(\omega^\star) \cdot \left(\frac{\partial^2 \gL(\omega, \theta^\star)}{\partial \omega^2}\right)^{-1} \cdot \frac{\partial^2 \gL(\omega, \theta^\star)}{\partial\omega\partial\theta}\Bigg|_{\omega^\star(\theta^\star)} 
\cdot \left(\frac{\partial^2 \predloss(\theta, \phi)}{\partial \theta^2}\right)^{-1} \cdot \frac{\partial^2 \predloss(\theta, \phi)}{\partial\theta\partial\phi}\Bigg|_{\theta^\star(\phi)}
  \end{aligned}
\end{equation}

\begin{figure}[t]
  \vspace*{-5mm}
  \hspace*{-2mm}
  \begin{minipage}[t]{0.4\textwidth}
    \begin{figure}[H]
      \centering
      \includegraphics[width=\textwidth]{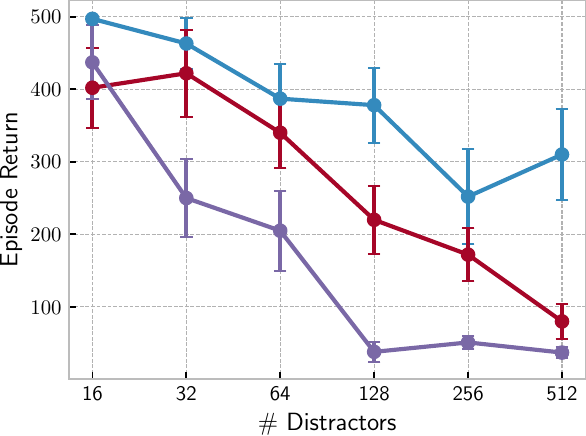}
      \cblock{52}{138}{189} \scriptsize TaskMet \hspace{1mm}
      \cblock{122}{104}{166} \scriptsize MLE \hspace{1mm}
      \cblock{166}{6}{40} OMD
      \caption{Results on the cartpole with distracting states
        \citep{nikishin2022control}.}
      \label{fig:distractors}
    \end{figure}
  \end{minipage}
  \hfill
  \begin{minipage}[t]{0.57\textwidth}
    \begin{figure}[H]
      \centering
      \resizebox{\textwidth}{!}{
        \begin{tikzpicture}
          \node[anchor=south west,inner sep=0] (image) at (0,0) {\includegraphics[width=3.5in]{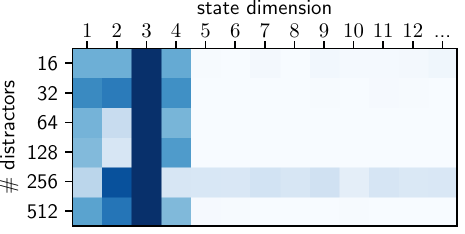}};
          \node[anchor=south east,inner sep=0,yshift=-14mm,xshift=-3mm] at (image.north east) {metric values {\color{gray}(darker=higher)}};
          \draw[black,ultra thick,rounded corners] (0.53in,1.45in) rectangle ++(.9in,4mm);
          \node at (1in,1.7in) {\color{gray} real states};
        \end{tikzpicture}
      }
      \caption{Our learned metric successfully distinguishes the real
        states from the distracting states, \ie
        the real states take a higher metric value.
      }
      \label{fig:learned-metric-distractors}
    \end{figure}
  \end{minipage}
\end{figure}

\begin{figure}[t]
  \centering
  \includegraphics[scale=0.7]{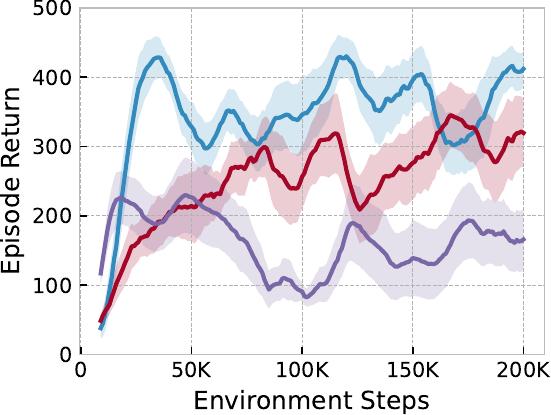}
  \hspace{5mm}
  \includegraphics[scale=0.7]{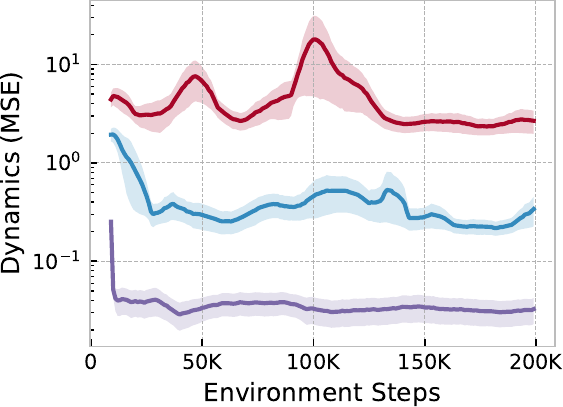} \\
  \cblock{52}{138}{189} TaskMet \hspace{2mm}
  \cblock{122}{104}{166} MLE \hspace{2mm}
  \cblock{166}{6}{40} OMD
  \caption{Results on cartpole with a reduced model capacity from \citet{nikishin2022control}.}
  \label{fig:model-capacity}
\end{figure}

\subsubsection{Experimental results}
We replicated experiments from \citet{nikishin2022control} on the
Cartpole environment. The first experiment involved state
distractions, where the state of the agent was augmented with noisy
and uninformative values. In this setting,
we considered an unconditional diagonal metric of dimension $n$, which
is the dimension of the state space, \ie
$\Lambda_\phi \defeq \text{diag}(\phi)$, where $\phi \in \mathbb{R}^n$.
As shown in \cref{fig:distractors}, the MLE method performed the worst across
different numbers of distracting states, as it allocated its capacity
to learn distracting states as well. TaskMet outperformed the other
methods in all scenarios. The superior performance of TaskMet with
distracting states can be attributed to the metric's ability to
explicitly distinguish informative states from noise states using the
task loss and then train the model using the given metric, as shown in
\cref{fig:learned-metric-distractors}.
The learned metric in \cref{fig:learned-metric-distractors} assigned
the highest weight to the third dimension of the state space, which
corresponds to the pole angle --- the most indicative dimension for
the reward. This shows that the metric can differentiate state
dimensions based on their importance to the task.

We also consider a setting with reduced model capacity, where the
network is under-parametrized, forcing the model to prioritize how it
allocates its capacity. In this scenario, we employ a full conditional
metric, denoted as $\Lambda_\phi = \Lambda_\phi(x)$, which enables the
metric to weigh dimensions and state-action pairs differently. We
conducted the experiment using a model size of 3 hidden units in the
layer. As depicted in \cref{fig:model-capacity}, TaskMet achieves a
better return on evaluation compared to MLE and OMD. Additionally, it
is evident that TaskMet achieves a lower MSE on the model predictions
compared to OMD, indicating that learning with the metric
contributes to a better dynamics model.

\section{Conclusion and discussion}

In conclusion, this paper addresses the challenge of combining task
and prediction losses in task-based model learning. While task-based
learning methods offer the advantage of discovering task-relevant
features and data samples without explicit inductive biases, the
current trend of using task loss alongside prediction loss has
potential limitations. These limitations include overfitting of the
prediction model to a specific task, rendering it ineffective for
other tasks, and the lack of interpretability in the task-relevant
features learned by the prediction model.

To overcome these limitations, the paper introduces the concept of
task-driven metric learning, which integrates the task loss into a
parameterized prediction loss. This approach enables end-to-end
learning of metrics to train prediction models, allowing the models to
focus on task-relevant features and dimensions in the prediction
space. Moreover, the resulting prediction models become more
interpretable, as metric learning serves as a preconditioning step for
gradient-based model training. The effectiveness of the method is
shown using different scales of experimental setting - decision
oriented tasks as well as downstream learning tasks.

One of the limitations of the method is stability of learning the metric.
Bad gradients can lead collapsed metric which can lead to
unrecoverable bad predictions. Hence, hyper-parameter tuning of
learning rate for metric learning and parameterization choices of the
metric are crucial.
Possible extensions to this work includes end-to-end metric learning
with multiple task losses, learning metric for training dynamics
models to be used for long-horizon planning tasks, etc.

\subsection*{Acknowledgments}
We would like to thank
Arman Zharmagambetov,
Brian Karrer,
Claas Voelcker,
Karen Ullrich,
Leon Bottou,
Maximilian Nickel, and
Mike Rabbat
insightful comments and discussions.

\bibliographystyle{plainnat}
\bibliography{refs}

\appendix
\newpage

\section{The implicit function theorem}
\label{sec:ift}
We used the implicit function theorem to compute the derivative
of the prediction model with respect to the metric's parameters
in \cref{eq:grad-phi}.
For completeness, this section briefly presents the standard
implicit function theorem, \cf \citet{dini1878analisi} and
\citet[Theorem 1B.1]{dontchev2009implicit}:

\begin{theorem}[Implicit Function Theorem]
Let $f: \mathbb{R}^n \times
\mathbb{R}^m \rightarrow \R^n$ be a continuous differentiable function,
and let $x^\star, y^\star$ be a point satisfying $f(x^\star, y^\star)=
0$. If the Jacobian $\frac{\partial f(x^\star, y^\star)}{\partial y}$
is non-singular, then there exists an open set around $(x^\star, y^\star)$ and
a unique continuously differentiable function $g$ such that $y^\star =
g(x^\star)$ and $f(x, g(x))=0$. Additionally, the following relation
holds:
\begin{equation}
\frac{\partial g(x)}{\partial x} = - \left(\frac{\partial f(x, y^\star)}{\partial y} \right)^{-1} \frac{\partial f(x, y^\star)}{\partial x} |_{y^\star = g(x)}
\end{equation}
\end{theorem}

\section{Implementation Details}
\label{sec:impl-details}

\subsection{Decision Oriented Model Learning}
We replicated our experiments using the codebase provided by
\citet{shah2022decisionfocused}, which can be found on
\href{https://github.com/sanketkshah/LODLs}{github}. To ensure
consistency, we used the same hyperparameters as mentioned in the code
or article for the baselines. Our metric learning pipeline was added
on top of their code, and thus we focused on tuning hyperparameters
related to metric learning. The metric is parameterized as
$\Lambda_\phi(x) = L_\phi(x) L_\phi^\top(x) + \epsilon_\phi
\mathbb{I}_{n \times n}$, where $\epsilon_\phi$ is a learnable
parameter that explicitly controls the amount of Euclidean metric in
the predicted metric. This helps ensure the stability of metric
learning. We initialize the parameters in such a way that the
predicted metric is close to the Euclidean metric.
For each outer loop of metric update, we perform $K$ inner updates
to train the predictor.
Following the methodology of \citet{shah2022decisionfocused},
we conducted 50 runs with different seeds for each of the experiments,
where each method was evaluated on 10 different datasets, with 5 different
seeds used for each dataset.

\begin{table}[H]
  \caption{Hyper-parameters for Decision Oriented Learning Experiments}
  \label{tab:hyper-params-lodl}
  \centering
  \begin{tabular}{r|l}
    \toprule
    Hyper-Parameter   &   Values    \\
    \midrule
    $\Lambda_\phi$ learning rate & $10^{-3}$   \\
    $\Lambda_\phi$ hidden layer sizes & \verb![200]! \\
    Warmup steps & $500$ \\
    Inner Iterations ($K$) & $100$  \\
    Implicit derivative batchsize & $10$ \\
    Implicit derivative solver &
      Conjugate gradient on the normal equations \detail{($5$ iterations)} \\
    \bottomrule
  \end{tabular}
\end{table}

\subsection{Model-Based Reinforcement Learning}
We consider the work of \citet{nikishin2022control} as the baseline
for replicating the experiments, and we build upon their source
code. Our metric learning is just one additional step to their
method. We adopt exact same hyperparameters as their for dynamics
learning and Action-Value function learning. We focus on exploring and
tuning the hyper-parameters specific to the metric learning component
of the method.

\begin{table}[H]
  \caption{Hyper-parameters for the CartPole experiments}
  \label{tab:hyper-params-mbrl}
  \centering
  \begin{tabular}{r|l}
    \toprule
    Hyper-Parameter   &   Values    \\
    \midrule
    $\Lambda_\phi$ learning rate & $10^{-3}$   \\
    $\Lambda_\phi$ hidden layer sizes & \verb![32, 32]! \\
    Warmup steps & $5000$ \\
    Inner iterations ($K$) & $1$  \\
    Implicit derivative batchsize & $256$ \\
    Implicit derivative solver & Conjugate gradient on the normal equations \detail{(10 iterations)} \\
    \bottomrule
  \end{tabular}
\end{table}

For the state distractor experiments, we parameterize the metric as an
unconditional diagonal matrix, denoted as
$\Lambda_\phi = \text{diag}(\phi)$ where $\phi \in \mathbb{R}^n$ and
$n$ is the dimension of the state space.
In addition, we consider a hyper-parameter of
\emph{metric parameterization}, for which we either
take normalize or unnormalized metric. When we refer to normalizing
the metric, we mean constraining the norm of the $\phi$ vector to be
equal to the L2 norm of an euclidean metric which is used by MSE
method. This constrains the family of learnable metrics. To achieve
this, we set $\phi \defeq \frac{\phi}{\| \phi \|_2} \sqrt{n}$,
ensuring $\| \phi \|_2 = \|\mathbb{I}_{n\times n} \|_2 = \sqrt{n}$. We
also used L1 regularization on the metric output, to induce sparsity
in the metric. We sweep over three values of the regularization
coefficient - $[0.0, 0.001, 0.1]$. We ran a sweep over the $6$
combinations of hyperparameters - $[\text{unnormalized, normalized}]
\times [0.0, 0.001,0.1]$ and choose the best hyper-parameter
combination for each of the experiment. All the number reported in the
experiments are calculated over $10$ random seeds.

Our metric learning approach uses two implicit gradient
steps. Firstly, we take the implicit derivative through action-value
network parameters, approximating the inverse hessian to the identity,
similar to \citet{nikishin2022control}. Secondly, for the step through
dynamics network parameters, we calculate the exact implicit
derivative.

\end{document}